\documentclass{multibody2015_full_paper_IMECH}
\usepackage{mathptmx}
\usepackage{graphicx}
\usepackage[bottom]{footmisc}
\usepackage{amsmath}
\usepackage{amsfonts}
\usepackage{amssymb}
\usepackage{enumitem}

\newtheorem{remark}{Remark}
\newtheorem{definition}{Definition}
\newtheorem{assumption}{Assumption}
\begin{document}

\begin{center}
\fontsize{14}{16}{\bf
    \LARGE Forward Dynamics of Variable Topology Mechanisms  \\ \vspace{0.4ex}
    -- The Case of Constraint Activation
  }
\end{center}


\begin{center}
{\normalsize {\ \textbf{{\ Andreas M\"uller} }} }
\end{center}


\begin{center}
\begin{tabular}{c}
Institute of Robotics, Johannes Kepler University \\ 
Altenbergerstr. 69, 4040 Linz, Austria, a.mueller@jku.at%
\end{tabular}
\end{center}

\section*{ABSTRACT}

Many mechanical systems exhibit changes in their kinematic topology altering
the mobility. Ideal contact is the best known cause, but also stiction and
controlled locking of parts of a mechanism lead to topology changes. The
latter is becoming an important issue in human-machine interaction.
Anticipating the dynamic behavior of variable topology mechanisms requires
solving a non-smooth dynamic problem. The core challenge is a physically
meaningful transition condition at the topology switching events. Such a
condition is presented in this paper. Two versions are reported, one using
projected motion equations in terms of redundant coordinates, and another
one using the Voronets equations in terms of minimal coordinates. Their
computational properties are discussed. Results are shown for joint locking
of a planar 3R mechanisms and a 6DOF industrial manipulator.

\textbf{Keywords:} Time integration, variable topology, non-smooth dynamics,
momentum conservation.

\section{Introduction}

Variable topology mechanisms (VTM) form a class of mechanisms that can
switch between different kinematic topologies, and thus change their
kinematic mobility and possibly their DOF. This property is known from
kinematotropic mechanisms \cite{galletifangella,RoeschelARK2000,Wohlhart1996}%
. The latter transit between motion modes via kinematic singularities while
keeping their kinematic topology. VTM on the other hand change their
mobility due to constraint switching. This can be caused by joint locking or
switching bilateral constraints (e.g. friction or contact activation). In
this paper it assumed that, except at the switching events, the constraints
are smooth and scleronomic. The VTM are then referred to as \emph{%
quasi-scleronomic VTM}.

Topology variations lead to discontinuous system trajectories. This is
problematic for the numerical simulation as well as for the solution of the
inverse dynamics problem within model-based control schemes. Unilateral
contacts in multibody systems are the prevailing source of topology changes,
and have been the topic of intensive research and the driving force towards
a theory for the non-smooth mechanics. An overview can be found in \cite%
{Leine2009,PfeifferGlocker1996,Pfeiffer2001}. In this context the modeling
of friction is crucial, and various formulations have been reported, e.g. 
\cite{Haug1986,Pereira1996,Piedboeuf2004,Trinkle1997}. Another cause for
non-smooth dynamics are discontinuous (bilateral) constraints, i.e.
switching between different constraints. These became relevant in the
context of molecular systems, where the original model is replaced by
lower-dimensional surrogate models. The topology change is merely due to a
change in the kinematics without the need to consider further constitutive
laws (as in case of frictional contact). The dynamics simulation of such
models with variable kinematic topology has been addressed in \cite%
{Chang,Mukherjee2007,Poursina2011,Poursina2011b}. A condition for momentum
conservation within numerical time stepping schemes was proposed.

Beyond molecular dynamics, there is a recent interest in this type of
discontinuous systems. Moreover, it accounts for situations with great
significance for robotics, and for human-machine interaction in particular.
An emergency stop is an excellent example. In such an event the joint brakes
are activated. This does not lead to an immediate halt. Rather the system is
initially moving according to the brakes' sliding friction. The individual
joints eventually lock and enter the stiction phase. This occurs at
different time instances for the different joints. Now, as part of the
safety considerations of the human-machine interaction it is crucial to
anticipate the trajectory of a robot in emergency situations.

Topology variations must also be resolved within the inverse dynamics
providing the feed-forward term in model-based control schemes. As an
example, in \cite{Simas} a method for obstacle-avoiding control of serial
manipulators was proposed, where a virtual joint is introduced whose motion
is in some sense perpendicular to the obstacle. This virtual joint is
introduced only when possible collision is detected, thus leads to a
variation of the topology.

From a computational point of view, the forward dynamics of VTM with
switching constraints requires additional compatibility conditions on the
velocity and momentum. Such conditions are presented in the following. The
paper introduces such conditions for the general case where multiple
constraints are successively activated.

\section{Configuration Space of Quasi-Scleronomic VTM}

The configuration space (c-space) of a holonomic quasi-scleronomic VTM is
the time dependent variety%
\vspace{-1ex}
\begin{equation}
V_{t}:=h^{-1}\left( \mathbf{q},t\right)
\end{equation}%
defined by a system of \emph{quasi-scleronomic }geometric\emph{\ constraints 
}$h\left( \mathbf{q},t\right) =\mathbf{0}$. The latter are piecewise defined
constraints of the form%
\vspace{-1ex}%
\begin{equation}
h\left( \mathbf{q},t\right) =\left\{ 
\begin{array}{ll}
h_{1}\left( \mathbf{q}\right) , & t\in \lbrack t_{0},t_{1}) \\ 
h_{2}\left( \mathbf{q}\right) , & t\in \lbrack t_{1},t_{2}) \\ 
\cdots &  \\ 
h_{i}\left( \mathbf{q}\right) , & t\in \lbrack t_{i-1},t_{i})%
\end{array}%
\right.
\end{equation}%
where the constraint switching happens at $t_{i}$. Each individual
constraint $h_{i}\left( \mathbf{q}\right) =\mathbf{0}$ corresponds to a
kinematic topology, and defines a variety $V_{i}:=h_{i}^{-1}\left( \mathbf{q}%
\right) $, where $V_{i}\cap V_{j}\neq \emptyset $, which is the \emph{%
c-space of the VTM at topology} $i$. Each c-space $V_{i}$ is stratified into
smooth manifolds of regular points and subvarieties of singular points. The
local DOF at $\mathbf{q}\in V_{i}$ is the highest dimension of submanifolds
passing through that point: $\delta _{\text{loc}}\left( \mathbf{q}\right)
:=\dim _{\mathbf{q}}V_{i}$. The differential (instantaneous) DOF is $\delta
_{\text{diff}}\left( \mathbf{q}\right) :=\dim \ker \mathbf{J}_{i}\left( 
\mathbf{q}\right) $, with $\mathbf{J}_{i}$ being the Jacobian of the
constraint mapping $h_{i}$. A configuration $\mathbf{q}$ is regular if and
only if $\delta _{\text{diff}}$ is constant in a neighborhood of $\mathbf{q}$
in $V$.

\begin{definition}
A topology change at time $t_{i}$ with configuration $\mathbf{q}_{0}\left(
t_{i}\right) \in V_{i}\cap V_{i+1}$ is called \emph{regular} if and only if $%
\mathbf{q}_{0}$ is a regular point of $V_{i}$ and of $V_{i+1}$. Otherwise it
is called \emph{singular}.
\end{definition}

A regular topology change is thus a change of local DOF without an
intermediate drop of the differential DOF. This is necessarily due to
contact constraints or (time dependent) locking. Apparently this is not
possible for kinematotropic mechanisms that must pass through singularities
in order to switch between different-dimensional manifolds.

In this paper a momentum consistent formulation for the forward dynamics is
presented for \emph{VTM undergoing regular topology changes due to the
activation of constraints}.

Hence the VTM does not encounter a singularity during the topology change.
This fact is exploited for the momentum consistent transition condition
applicable within time stepping schemes. The method is tailored to the case
that additional constraints are activated.

\section{Suitable Forms of Motion Equations}

\subsection{Lagrange Equations of First Kind}

In the following a form of the equations of motions (EOM) is recalled that
will be suited for the derivation of the momentum balance. The starting
point are the EOM of an unconstrained MBS of the form $\mathbf{M}\left( 
\mathbf{q}\right) {\ddot{\mathbf{q}}}+\mathbf{C}\left( {\dot{\mathbf{q}},}%
\mathbf{q}\right) {\dot{\mathbf{q}}}+\mathbf{P}\left( {\mathbf{q}}\right) +%
\mathbf{Q}\left( {\dot{\mathbf{q}},\mathbf{q},t}\right) =\mathbf{u}\left( {t}%
\right) $. Here $\mathbf{M}$ is the generalized mass matrix, $\mathbf{C}{%
\dot{\mathbf{q}}}$ represents forces quadratic in velocity, $\mathbf{P}$
potential forces, and $\mathbf{Q}$ any other generalized forces inherent to
the system. The vector $\mathbf{u}$ represents the applied generalized
control forces. The vector $\mathbf{q}\in {\mathbb{V}}^{n}$ comprises $n$
generalized coordinates, where the variety ${\mathbb{V}}^{n}={\mathbb{R}}%
^{n_{p}}\times {\mathbb{T}}^{n_{r}}$ accounts for $n_{p}$ translational and $%
n_{r}$ rotational coordinates. It is assumed that $\mathbf{C}$ and $\mathbf{P%
}$ are smooth mappings. The reason for separating the generalized forces
that are neither potential forces nor quadratic in ${\dot{\mathbf{q}}}$ will
become clear in section \ref{secBlance}.

Assume that the MBS is subjected to a system of $m$ scleronomic geometric
constraints $h\left( \mathbf{q}\right) =\mathbf{0}$ that define the c-space $%
V=h^{-1}\left( \mathbf{0}\right) $. The constraint Jacobian, in the
corresponding velocity constraints $\mathbf{J}\left( \mathbf{q}\right) {\dot{%
\mathbf{q}}}$ $=\mathbf{0}$, leads to the Lagrangian EOM%
\begin{equation}
\mathbf{M}\left( \mathbf{q}\right) {\ddot{\mathbf{q}}}+\mathbf{C}\left( {%
\dot{\mathbf{q}},}\mathbf{q}\right) {\dot{\mathbf{q}}}+\mathbf{P}\left( {%
\mathbf{q}}\right) +\mathbf{Q}\left( {\dot{\mathbf{q}},\mathbf{q},t}\right) +%
\mathbf{J}^{T}\left( \mathbf{q},t\right) \mathbf{\lambda }=\mathbf{u}\left(
t\right)  \label{Lagrange}
\end{equation}%
The above geometric constraints give rise to the acceleration constraints 
\begin{equation}
\mathbf{J}\left( \mathbf{q}\right) {\ddot{\mathbf{q}}+\dot{\mathbf{J}}\left( 
\mathbf{q},{\dot{\mathbf{q}}}\right) \dot{\mathbf{q}}}=\mathbf{0}.
\label{accConstr}
\end{equation}%
For numerical treatment, frequently (\ref{Lagrange}) is combined with the
acceleration constraints to yield the index 1 DAE system%
\begin{equation}
\left( 
\begin{array}{cc}
\mathbf{M}\left( \mathbf{q}\right) & {\mathbf{J}}^{T}\left( \mathbf{q}\right)
\\ 
{\mathbf{J}}\left( \mathbf{q}\right) & \mathbf{0}%
\end{array}%
\right) \left( 
\begin{array}{c}
{\ddot{\mathbf{q}}} \\ 
\mathbf{\lambda }%
\end{array}%
\right) =\left( 
\begin{array}{l}
-\mathbf{C}\left( {\dot{\mathbf{q}},}\mathbf{q}\right) {\dot{\mathbf{q}}-%
\mathbf{P}\left( {\mathbf{q}}\right) -}\mathbf{Q}\left( {\dot{\mathbf{q}},%
\mathbf{q},t}\right) +\mathbf{u}\left( {t}\right) \\ 
-{\dot{\mathbf{J}}\left( \mathbf{q},{\dot{\mathbf{q}}}\right) \dot{\mathbf{q}%
}}%
\end{array}%
\right) .  \label{index1}
\end{equation}%
The advantage of this formulation is that the coefficient matrix is usually
sparse since the constraint Jacobian is usually sparse. The sparsity of the
mass matrix depends on how the unconstrained system is modeled. If the
latter is modeled in terms of absolute coordinates, $\mathbf{M}$ is
bock-diagonal. Otherwise, e.g. using relative coordinates, this is a dense
block-triangular matrix. Then the size $\left( n+m\right) \times \left(
n+m\right) $ of the coefficient matrix can become a burden for numerical
solution. Moreover, when the number of constraints is not constant, the size
of the coefficient matrix changes. A formulation with coefficient matrix of
constant dimension is presented in the next section. The formulation in
terms of minimal coordinates is recalled in section \ref{secVoronets}.

\subsection{Projected Motion Equations in terms of Redundant Coordinates
--Gau\ss ' Principle}

A system of EOM for the constrained system in terms of the complete set of
the (redundant) coordinates $\mathbf{q}\in {\mathbb{V}}^{n}$ can be derived
from Gau\ss '\ principle written in the form%
\begin{equation}
\left\{ 
\begin{array}{c}
\frac{1}{2}\left( {\ddot{\mathbf{q}}}-\mathbf{a}\right) ^{T}\mathbf{M}\left( 
\mathbf{q}\right) \left( {\ddot{\mathbf{q}}}-\mathbf{a}\right) \rightarrow
\min \\ 
\mathbf{J}\left( \mathbf{q}\right) {\ddot{\mathbf{q}}+\dot{\mathbf{J}}\dot{%
\mathbf{q}}}=\mathbf{0}%
\end{array}%
\right\}  \label{Gauss}
\end{equation}%
where $\mathbf{a}=\mathbf{M}^{-1}(\mathbf{u}-\mathbf{C}{\dot{\mathbf{q}}-%
\mathbf{P}-}\mathbf{Q})$ is the acceleration of the unconstrained system.
Assuming non-redundant constraints, the solution ${\ddot{\mathbf{q}}}$
solving (\ref{Gauss}) is%
\begin{equation}
{\ddot{\mathbf{q}}=}\mathbf{M}^{-1}\left( \mathbf{q}\right) \mathbf{N}_{%
\mathbf{J},\mathbf{M}}^{T}%
\big%
(\mathbf{u}\left( {\mathbf{q},t}\right) -\mathbf{C}\left( {\dot{\mathbf{q}},}%
\mathbf{q}\right) {\dot{\mathbf{q}}-\mathbf{P}\left( {\mathbf{q}}\right) -}%
\mathbf{Q}\left( {\dot{\mathbf{q}},\mathbf{q},t}\right) 
\big%
)-\mathbf{J}_{\mathbf{M}}^{+}\left( \mathbf{q}\right) {\dot{\mathbf{J}}%
\left( \mathbf{q},{\dot{\mathbf{q}}}\right) \dot{\mathbf{q}}}  \label{eq1}
\end{equation}%
with $\mathbf{J}_{\mathbf{M}}^{+}=\mathbf{M}^{-1}\mathbf{J}^{T}\left( 
\mathbf{JM}^{-1}\mathbf{J}^{T}\right) ^{-1}\,$ being the $\mathbf{M}$%
-weighted right pseudoinverse of $\mathbf{J}$, and $\mathbf{N}_{\mathbf{J},%
\mathbf{M}}=\mathbf{I}-\mathbf{J}_{\mathbf{M}}^{+}\mathbf{J}$ the
corresponding projector to the null-space of $\mathbf{J}$. This formulation
is free of Lagrange multipliers and further does not require selection of
independent (minimal) coordinates, as minimal coordinate formulations for
constrained MBS do (next section).

As reported in \cite{NODY}, using ${\dot{\mathbf{J}}\dot{\mathbf{q}}}=-%
\mathbf{J}{\ddot{\mathbf{q}}}$, the equations (\ref{eq1}) can be written%
\begin{equation}
\mathbf{N}_{\mathbf{J},\mathbf{M}}^{T}%
\big%
(\mathbf{M}\left( \mathbf{q}\right) {\ddot{\mathbf{q}}}+\mathbf{C}\left( {%
\dot{\mathbf{q}},}\mathbf{q}\right) {\dot{\mathbf{q}}+\mathbf{P}\left( {%
\mathbf{q}}\right) +}\mathbf{Q}\left( {\dot{\mathbf{q}},\mathbf{q},t}\right)
-\mathbf{u}\left( {\mathbf{q},t}\right) 
\big%
)=\mathbf{0}.  \label{projEOM}
\end{equation}

The formulation (\ref{projEOM}) is advantageous since the unaltered EOM of
the unconstrained system are simply projected onto the c-space (actually its
cotangent space). Moreover, a change in the constraints only affects the
null-space projector respectively the pseudoinverse.

\begin{remark}
The motion equations (\ref{eq1}) form a system of $n$ independent ODEs in $%
\mathbf{q}\left( t\right) $ that can be used for forward dynamics analysis.
The system (\ref{projEOM}), on the other hand, consists of $n$ equations of
which only $n-m$ are independent, since $\mathrm{rank}~\mathbf{N}_{\mathbf{J}%
,\mathbf{M}}=n-m$. It is thus not directly applicable to forward dynamics
simulation, but allows for a very efficient inverse dynamics solution
applicable for model-based control schemes in redundant coordinates \cite%
{MuellerHufnagel2012}. Nevertheless, the equations (\ref{projEOM}) together
with (\ref{accConstr}) form a system of $n+m$ motion equations of which $n$
are independent.
\end{remark}

\begin{remark}
The image space of the weighted null-space projector $\mathbf{N}_{\mathbf{J},%
\mathbf{M}}^{T}$ depends on the weight $\mathbf{M}$. This arising from the
solution of (\ref{Gauss}), respectively the elimination of the Lagrange
multipliers, in order to ensure the dynamic consistency of the projection in
(\ref{eq1}). On the other hand, in the projected motion equations (\ref%
{projEOM}) any projector to the null-space of $\mathbf{J}$ can be used. A
particularly simple choice is the non-weighted pseudoinverse $\mathbf{J}%
^{+}:=\mathbf{J}^{T}\left( \mathbf{JJ}^{T}\right) ^{-1}$ with corresponding $%
\mathbf{N}_{\mathbf{J}}\equiv \mathbf{N}_{\mathbf{J},\mathbf{I}}=\mathbf{I}-%
\mathbf{J}^{+}\mathbf{J}$. Then it is $\mathbf{N}_{\mathbf{J}}=\mathbf{N}_{%
\mathbf{J}}^{T}$, and thus%
\begin{equation}
\mathbf{N}_{\mathbf{J}}%
\big%
(\mathbf{M}\left( \mathbf{q}\right) {\ddot{\mathbf{q}}}+\mathbf{C}\left( {%
\dot{\mathbf{q}},}\mathbf{q}\right) {\dot{\mathbf{q}}+\mathbf{P}\left( {%
\mathbf{q}}\right) +}\mathbf{Q}\left( {\dot{\mathbf{q}},\mathbf{q},t}\right)
-\mathbf{u}\left( {\mathbf{q},t}\right) 
\big%
)=\mathbf{0}.  \label{projEOM2}
\end{equation}
\end{remark}

\subsection{Motion Equations in terms of Minimal Coordinates --Voronets
Equations%
\label{secVoronets}%
}

Due to the $m$ constraints, only $n-m$ coordinates are independent (assuming
non-redundant constraints). In contrast to the above formulation, the
selection of a minimal set of independent generalized coordinates is not
possible globally, but only feasible locally at a given configuration $%
\mathbf{q}$. Assume that the c-space $V$ is a smooth manifold at the
considered configuration, i.e. $\delta _{\text{loc}}\left( \mathbf{q}\right)
=n-m$, and $n-m$ of the coordinates can be used to parameterize the MBS
configuration. Denote with $\mathbf{s}\in {\mathbb{V}}^{n-m}$ the vector of
(locally) independent minimal coordinates. The coordinate vector can be
rearranged as $\mathbf{q}=\left( \mathbf{p},\mathbf{s}\right) $, where $%
\mathbf{p}$ is the vector of $m$ dependent coordinates, and the Jacobian
reads accordingly $\mathbf{J}=\left( \mathbf{J}_{\mathbf{p}},\mathbf{J}_{%
\mathbf{s}}\right) $. Then an orthogonal complement for the constraint
Jacobian is given explicitly as%
\begin{equation}
\mathbf{F}=\left( 
\begin{array}{c}
-\mathbf{J}_{\mathbf{p}}^{-1}\mathbf{J}_{\mathbf{s}} \\ 
\mathbf{I}_{n-m}%
\end{array}%
\right) .  \label{F}
\end{equation}%
That is, ${\dot{\mathbf{q}}}=\mathbf{F}{\dot{\mathbf{s}}}$ and $\mathbf{J}{%
\mathbf{F}}\equiv \mathbf{0}$. Premultiplication of (\ref{Lagrange}) with $%
\mathbf{F}^{T}$\textbf{\ }yields a system of $n-m$ equations%
\begin{equation}
\mathbf{F}^{T}\left( \mathbf{q}\right) 
\big%
(\mathbf{M}\left( \mathbf{q}\right) {\ddot{\mathbf{q}}}+\mathbf{C}\left( {%
\dot{\mathbf{q}},}\mathbf{q}\right) {\dot{\mathbf{q}}+\mathbf{P}\left( {%
\mathbf{q}}\right) +}\mathbf{Q}\left( {\dot{\mathbf{q}},\mathbf{q},t}\right)
-\mathbf{u}\left( {\mathbf{q},t}\right) 
\big%
)=\mathbf{0}.
\end{equation}%
Together with (\ref{accConstr}) this forms a system of $n$ independent
motions equations. This can be reduced to the minimal coordinates by
replacing ${\ddot{\mathbf{q}}}=\mathbf{F}{\ddot{\mathbf{s}}}+{\dot{\mathbf{F}%
}\dot{\mathbf{s}}}$, which yields the Voronets equations%
\begin{equation}
\overline{\mathbf{M}}\left( \mathbf{q}\right) {\ddot{\mathbf{s}}}+\overline{%
\mathbf{C}}\left( {\dot{\mathbf{s}},}\mathbf{q}\right) {\dot{\mathbf{s}}+{%
\overline{\mathbf{P}}}\left( {\mathbf{q}}\right) +\overline{\mathbf{Q}}}%
\left( {\dot{\mathbf{s}},\mathbf{q},t}\right) -\overline{\mathbf{u}}\left( {%
\mathbf{q},t}\right) =\mathbf{0}  \label{Voronets}
\end{equation}%
with%
\vspace{-2ex}%
\begin{equation}
\overline{\mathbf{M}}:=\mathbf{F}^{T}\mathbf{MF,\ \overline{\mathbf{C}}:=F}%
^{T}\left( \mathbf{CF}+\mathbf{M}{\dot{\mathbf{F}}}\right) ,\ \overline{%
\mathbf{P}}:=\mathbf{F}^{T}\mathbf{P,\ }\overline{\mathbf{Q}}:=\mathbf{F}^{T}%
\mathbf{Q},\ \overline{\mathbf{u}}:=\mathbf{F}^{T}\mathbf{u}.
\end{equation}%
This is a system of $n-m$ second order ODEs in the $n-m$ independent
coordinates $\mathbf{s}$. $\overline{\mathbf{M}}$ is the mass matrix
projected on the subspace of minimal coordinates. The coordinate formulation
(\ref{Voronets}) goes back to Voronets \cite{voronets1901}, which is a
special case of the Maggi equations \cite{Maggi1901} in holonomic
coordinates.

The problematic point with any minimal coordinate formulation is that a
chosen set of independent coordinates (determining the orthogonal
complement) is only valid within a certain part of the c-space. A set of
coordinates fails as local parameters when the matrix $\mathbf{J}_{\mathbf{p}%
}$ becomes singular. Throughout this paper it assumed that no constraint
singularities occur and that a valid set of minimal coordinates has been
chosen.

\section{Compatibility Condition for general Switching Constraints}

\subsection{General Form of Switching Constraints}

Without loss of generality a constraint switching at time $t=0$ is
considered, which can be formulated as%
\begin{equation}
h\left( \mathbf{q},t\right) =\left\{ 
\begin{array}{cc}
h_{-}\left( \mathbf{q}\right) , & t<0 \\ 
h_{+}\left( \mathbf{q}\right) , & t\geq 0%
\end{array}%
\right. ,\ \ \ \ \mathbf{J}\left( \mathbf{q},t\right) =\left\{ 
\begin{array}{cc}
\mathbf{J}_{-}\left( \mathbf{q}\right) , & t<0 \\ 
\mathbf{J}_{+}\left( \mathbf{q}\right) , & t\geq 0%
\end{array}%
\right.  \label{switch}
\end{equation}%
where $h_{-}$ and $h_{+}$ are respectively the constraints before and after
the switching event, and $\mathbf{J}_{-}$ and $\mathbf{J}_{+}$ are the
corresponding constraint Jacobians. Denote the switch configuration with $%
\mathbf{q}_{0}:=\mathbf{q}\left( 0\right) $. It is assumed that the
individual constraints $h_{-}$ and $h_{+}$ are smooth at $t=0$.

\subsection{Kinematic Compatibility}

The velocity before the switching event is denoted with ${\dot{\mathbf{q}}}%
_{-}$, and the one after the event with ${\dot{\mathbf{q}}}_{+}$. The
velocity constraints are satisfied at any time, i.e.%
\begin{eqnarray}
{\mathbf{J}}_{+}\left( \mathbf{q}_{0}\right) {\dot{\mathbf{q}}}_{+} &=&%
\mathbf{0}  \label{velConstraint1} \\
{\mathbf{J}}_{-}\left( \mathbf{q}_{0}\right) {\dot{\mathbf{q}}}_{-} &=&%
\mathbf{0}.  \label{velConstraint2}
\end{eqnarray}%
These two conditions can be combined to the following condition on the
velocity jump%
\begin{equation}
{\mathbf{J}}_{+}\left( \mathbf{q}_{0}\right) {\Delta \dot{\mathbf{q}}+{%
\mathbf{J}}_{+}\left( \mathbf{q}_{0}\right) \dot{\mathbf{q}}}_{-}=\mathbf{0}.
\label{kinCompatibility}
\end{equation}

\subsection{Momentum Balance%
\label{secBlance}%
}

The switching causes impulsive constrain forces, and thus a discontinuous
change of the generalized momentum. This cannot be captured by the projected
EOM (\ref{projEOM}) or (\ref{Voronets}) since one particular set of
constraints is incorporated in order to eliminate the Lagrange multipliers.
Starting instead from the unconstrained equations, imposing the
quasi-scleronomic constraints leads to the Lagrange equation%
\begin{equation}
\mathbf{M}\left( \mathbf{q}\right) {\ddot{\mathbf{q}}}+\mathbf{C}\left( {%
\dot{\mathbf{q}},}\mathbf{q}\right) {\dot{\mathbf{q}}+\mathbf{P}\left( {%
\mathbf{q}}\right) +}\mathbf{Q}\left( {\dot{\mathbf{q}},\mathbf{q},t}\right)
+\mathbf{J}^{T}\left( \mathbf{q},t\right) \mathbf{\lambda }=\mathbf{u}\left( 
\mathbf{q},t\right) .  \label{Lagrange2}
\end{equation}%
The switching causes a jump in the velocity and thus in the generalized
momentum due to the constraint Jacobian in (\ref{switch}). It is assumed
that ${\mathbf{J}}_{+}$ and ${\mathbf{J}}_{-}$ are smooth for $t\in \left(
-\varepsilon ,0\right) $ and $t\in \left[ 0,\varepsilon \right) $,
respectively. The momentum conservation at the switching time $t=0$, with
corresponding configuration $\mathbf{q}_{0}=\mathbf{q}\left( 0\right) $,
leads to%
\begin{equation}
\int_{0}^{\varepsilon }\left( \mathbf{M}\left( \mathbf{q}\right) {\ddot{%
\mathbf{q}}}+\mathbf{C}\left( {\dot{\mathbf{q}},}\mathbf{q}\right) {\dot{%
\mathbf{q}}+\mathbf{P}\left( {\mathbf{q}}\right) +}\mathbf{Q}\left( {\dot{%
\mathbf{q}},\mathbf{q},t}\right) +\mathbf{J}^{T}\left( \mathbf{q},t\right) 
\mathbf{\lambda }\right) dt=\int_{0}^{\varepsilon }\mathbf{u}\left( \mathbf{q%
},t\right) dt
\end{equation}%
for $\varepsilon \rightarrow 0$. Noting that $\mathbf{M}$ is constant during
the switching, and that $\mathbf{J}_{+}^{T}$ is smooth at $t\in \left[
0,\varepsilon \right) $ yields the relation%
\begin{equation}
\mathbf{M}\left( \mathbf{q}_{0}\right) {\Delta \dot{\mathbf{q}}}%
+\int_{0}^{\varepsilon }\left( \mathbf{C}\left( {\dot{\mathbf{q}},}\mathbf{q}%
\right) {\dot{\mathbf{q}}+\mathbf{P}\left( {\mathbf{q}}\right) +\mathbf{Q}}%
\left( {\dot{\mathbf{q}},\mathbf{q},t}\right) \right) dt+\ \mathbf{J}%
_{+}^{T}\left( \mathbf{q}_{0}\right) \Lambda =\mathbf{U}\left( \mathbf{q}%
_{0}\right)
\end{equation}%
on the velocity jump ${\Delta \dot{\mathbf{q}}:=\dot{\mathbf{q}}}_{+}-{\dot{%
\mathbf{q}}}_{-}$\textbf{. }Here $\Lambda :=\int_{0}^{\varepsilon }\mathbf{%
\lambda }dt$ is the impulsive constraint force due to the constraint
switching, and $\mathbf{U}\left( \mathbf{q}_{0}\right)
=\int_{0}^{\varepsilon }\mathbf{u}\left( \mathbf{q},t\right) dt$. The
integral over ${\mathbf{P}}$ vanishes for $\varepsilon \rightarrow 0$ since
the potential forces depend smoothly on $\mathbf{q}$. The vector of Coriolis
and centrifugal forces $\mathbf{C}\left( {\dot{\mathbf{q}},}\mathbf{q}%
\right) {\dot{\mathbf{q}}}$ is a quadratic form in ${\dot{\mathbf{q}}}$. The 
$i$th component of this force reads explicitly $\sum_{j,k}{\Gamma }%
_{ijk}\left( \mathbf{q}\right) \dot{q}^{j}\dot{q}^{k}$, with ${\Gamma }%
_{ijk} $ being the Christoffel symbols corresponding to the mass matrix.
Although the velocities are discontinuous during the event, they are
bounded. Hence also the integral over the quadratic form $\mathbf{C}\left( {%
\dot{\mathbf{q}},}\mathbf{q}\right) {\dot{\mathbf{q}}}$ vanishes for $%
\varepsilon \rightarrow 0$. The only remaining integral term is that for ${%
\mathbf{Q}}$ representing general velocity dependent forces, such as
friction.

\begin{assumption}
\label{assumption1}%
The contribution of the forces $\mathbf{Q}$ to the momentum balance can be
neglected.
\end{assumption}

With this presumption the momentum balance becomes%
\begin{equation}
\mathbf{M}\left( \mathbf{q}_{0}\right) {\Delta \dot{\mathbf{q}}}+\mathbf{J}%
_{+}^{T}\left( \mathbf{q}_{0}\right) \Lambda =\mathbf{U}\left( \mathbf{q}%
_{0}\right) .
\end{equation}

\subsection{Overall Transition Condition}

Combining the momentum balance with the kinematic conditions (\ref%
{kinCompatibility}) gives rise to the following system%
\begin{equation}
\left( 
\begin{array}{cc}
\mathbf{M}\left( \mathbf{q}_{0}\right) & {\mathbf{J}}_{+}^{T}\left( \mathbf{q%
}_{0}\right) \\ 
{\mathbf{J}}_{+}\left( \mathbf{q}_{0}\right) & \mathbf{0}%
\end{array}%
\right) \left( 
\begin{array}{c}
{\Delta \dot{\mathbf{q}}} \\ 
\Lambda%
\end{array}%
\right) =\left( 
\begin{array}{c}
\mathbf{U}\left( \mathbf{q}_{0}\right) \\ 
-{\mathbf{J}}_{+}\left( \mathbf{q}_{0}\right) {\dot{\mathbf{q}}}_{-}%
\end{array}%
\right) .  \label{sys1}
\end{equation}%
A solution ${\Delta \dot{\mathbf{q}}}$ of (\ref{sys1}) is an admissible
change of generalized velocity such that the constraints prior and after the
switching event as well as the momentum balance are satisfied. This system
of $n+m_{+}$ equations for the $n+m_{+}$ unknowns ${\dot{\mathbf{q}}}_{+}$
and $\Lambda $, has a unique solution provided that the constraints are
non-redundant.

\subsection{Incorporation into a Time Integration Scheme}

The transition condition (\ref{sys1}) is only required at a switching event.
Any motion equations (such as (\ref{Lagrange}), (\ref{eq1}) or (\ref%
{Voronets})) can be numerically integrated until the event, providing the
velocity ${\dot{\mathbf{q}}}_{-}$ prior to the event and the configuration $%
\mathbf{q}_{0}\in V$ satisfying the geometric constraints. The velocity jump
at $t=0$ could then be determined by solving (\ref{sys1}). The compatibility
condition (\ref{sys1}) yields the full (redundant) state of the system, from
which any desired set of coordinates can be selected, e.g. minimal
coordinates.

The coefficient matrix in (\ref{sys1}) is exactly the one in the index 1
formulation (\ref{index1}) evaluated at $\mathbf{q}_{0}$. Hence the system (%
\ref{sys1}) can be solved in parallel with the EOM if the EOM (\ref{index1})
is integrated, However, for large $n$ and a large number $m$ of constraints
(e.g. switching molecular systems \cite{Poursina2011b}) the EOM in redundant
or minimal coordinates may be computationally simpler. Then the condition (%
\ref{sys1}) is still applicable, but would require solving a system of $%
n+m_{+}$ equations. To avoid this, a transition condition for either
formulation is required. This can be derived for the special case of
activation of additional constraint, as shown in the following.

The impulsive force $\mathbf{U}\left( \mathbf{q}_{0}\right) $ can possibly
be substituted by a contact model.

\vspace{2ex}%

\section{Compatibility Condition for Activation of additional Constraints}

\subsection{Activation of additional Constraints}

A frequently encountered type of VTM switching constraints is that \emph{%
further constraints are activated in addition to constraints already active
before the switching event}. This can be modeled by the following form of
the quasi-scleronomic constraints%
\begin{equation}
h\left( \mathbf{q},t\right) =\left\{ 
\begin{array}{ll}
h_{-}\left( \mathbf{q}\right) :=h_{1}\left( \mathbf{q}\right) , & t<0 \\ 
& 
\vspace{-1.6ex}
\\ 
h_{+}\left( \mathbf{q}\right) :=\left( 
\begin{array}{c}
h_{1}\left( \mathbf{q}\right) \\ 
h_{2}\left( \mathbf{q}\right)%
\end{array}%
\right) , & t\geq 0%
\end{array}%
\right. ,\ \ \ \ \mathbf{J}\left( \mathbf{q},t\right) =\left\{ 
\begin{array}{ll}
\mathbf{J}_{-}\left( \mathbf{q}\right) :=\mathbf{J}_{1}\left( \mathbf{q}%
\right) , & t<0 \\ 
& 
\vspace{-1.6ex}
\\ 
\mathbf{J}_{+}\left( \mathbf{q}\right) :=\left( 
\begin{array}{c}
\mathbf{J}_{1}\left( \mathbf{q}\right) \\ 
\mathbf{J}_{2}\left( \mathbf{q}\right)%
\end{array}%
\right) , & t\geq 0\ \ .%
\end{array}%
\right.  \label{switchConstr}
\end{equation}%
Here $h_{1}\left( \mathbf{q}\right) =\mathbf{0}$ is a set of $m_{1}$
persistent constraints throughout the switching to which another set of $%
m_{2}$ constraints $h_{2}\left( \mathbf{q}\right) =\mathbf{0}$ is added at $%
t=0$.

The c-space before the switching event is $V_{-}:=h_{-}^{-1}\left( \mathbf{0}%
\right) $ and that after is $V_{+}:=h_{+}^{-1}\left( \mathbf{0}\right) $.
Due to the persistent constraints it is $V_{-}:=h_{-}^{-1}\left( \mathbf{0}%
\right) =h_{1}^{-1}\left( \mathbf{0}\right) $ and $V_{+}:=h_{+}^{-1}\left( 
\mathbf{0}\right) =h_{1}^{-1}\left( \mathbf{0}\right) \cap h_{2}^{-1}\left( 
\mathbf{0}\right) $, and thus $V_{+}\subset V_{-}:=h_{1}^{-1}\left( \mathbf{0%
}\right) $.

Denote with $\mathbf{q}_{0}:=\mathbf{q}\left( 0\right) $ the configuration
where the switching takes place at $t=0$.

\begin{assumption}
\label{assumption2}%
Both constraint mappings, $h_{1}$ and $h_{2}$, are smooth at $\mathbf{q}_{0}$%
. That is, $V_{1}=h_{1}^{-1}\left( \mathbf{0}\right) $ and $%
V_{2}=h_{2}^{-1}\left( \mathbf{0}\right) $ are locally smooth manifolds at $%
\mathbf{q}_{0}$. Further each $\mathbf{J}_{1}\left( \mathbf{q}_{0}\right) $
and $\mathbf{J}_{2}\left( \mathbf{q}_{0}\right) $ has full rank $m_{1}$ and $%
m_{2}$, respectively. That is, the individual constraint systems are
non-redundant.
\end{assumption}

\begin{assumption}
\label{assumption3}%
The constraints $h_{1}$ and $h_{2}$ are locally algebraically independent.
That is, $V_{+}=V_{1}\cap V_{2}$, with $V_{1}=h_{1}^{-1}\left( \mathbf{0}%
\right) $ and $V_{2}=h_{2}^{-1}\left( \mathbf{0}\right) $, is a smooth
manifold of dimension $n-m_{1}-m_{2}$. That is, the overall constraint
system is non-redundant.
\end{assumption}

\vspace{2ex}%

\subsection{Transition Condition for the Lagrangian Motion Equations}

With the partitioned form of the constraints (\ref{switchConstr}), the
transition conditions (\ref{index1}) are 
\begin{equation}
\left( 
\begin{array}{ccc}
\mathbf{M}\left( \mathbf{q}_{0}\right) & {\mathbf{J}}_{1}^{T}\left( \mathbf{q%
}_{0}\right) & {\mathbf{J}}_{2}^{T}\left( \mathbf{q}_{0}\right) \\ 
{\mathbf{J}}_{1}\left( \mathbf{q}_{0}\right) & \mathbf{0} & \mathbf{0} \\ 
{\mathbf{J}}_{2}\left( \mathbf{q}_{0}\right) & \mathbf{0} & \mathbf{0}%
\end{array}%
\right) \left( 
\begin{array}{c}
{\Delta \dot{\mathbf{q}}} \\ 
\Lambda _{1} \\ 
\Lambda _{2}%
\end{array}%
\right) =\left( 
\begin{array}{c}
\mathbf{U}\left( \mathbf{q}_{0}\right) \\ 
-{\mathbf{J}}_{1}\left( \mathbf{q}_{0}\right) {\dot{\mathbf{q}}}_{-} \\ 
-{\mathbf{J}}_{2}\left( \mathbf{q}_{0}\right) {\dot{\mathbf{q}}}_{-}%
\end{array}%
\right)  \label{trans1}
\end{equation}%
where $\Lambda _{1}$ and $\Lambda _{2}$ are the impulsive constraint forces
due to the persistent and additional constraints, respectively.

\begin{itemize}
\item Solving the system (\ref{trans1}) requires inverting the $\left(
n+m_{1}+m_{2}\right) \times \left( n+m_{1}+m_{2}\right) $ matrix. The mass
matrix $\mathbf{M}$ in (\ref{trans1}) can be inverted with the same method
as for the solution of the system (\ref{index1}) within the time integration
scheme. For instance, if the EOM are evaluated with an $O\left( n\right) $
method, then also the system (\ref{trans1}) can be evaluated with this
method.

\item The condition can be used in conjunction with any dynamics formulation
(redundant or minimal coordinates). The compatibility condition (\ref{trans1}%
) yields the full (redundant) state of the system, from which any desired
set of coordinates can be selected, e.g. minimal coordinates. The system (%
\ref{trans1}) is only solved at the switching event.

\item If the constraints $h_{1}\left( \mathbf{q}\right) =\mathbf{0}$ and $%
h_{2}\left( \mathbf{q}\right) =\mathbf{0}$ are independent (assumption \ref%
{assumption3}), and $\mathbf{J}_{1}$ and $\mathbf{J}_{2}$ have full rank
(assumption \ref{assumption2}), the coefficient matrix in (\ref{trans1}) has
full rank $n+m_{1}+m_{2}$.
\end{itemize}

\subsection{Transition Condition in Terms of the full Set of Coordinates}

\subsubsection{Momentum Balance}

In the projected EOM (\ref{projEOM}), the Lagrange multipliers are
eliminated so that the above approach is not applicable directly.
Nevertheless the constraint forces can be retained if the projection in the
equations (\ref{projEOM}) is restricted to the set of persistent constraints 
$h_{1}\left( \mathbf{q}\right) =\mathbf{0}$. Imposing the additional
constraints leads to corresponding constraint forces.

The projected EOM accounting for the persistent constraints are%
\begin{equation}
\mathbf{N}_{\mathbf{J}_{1},\mathbf{M}}^{T}%
\big%
(\mathbf{M}\left( \mathbf{q}\right) {\ddot{\mathbf{q}}}+\mathbf{C}\left( {%
\dot{\mathbf{q}},}\mathbf{q}\right) {\dot{\mathbf{q}}+{\mathbf{P}\left( {%
\mathbf{q}}\right) }+}\mathbf{Q}\left( {\dot{\mathbf{q}},\mathbf{q},t}%
\right) +\mathbf{J}^{T}\left( \mathbf{q},t\right) \mathbf{\lambda }-\mathbf{u%
}\left( {\mathbf{q},t}\right) 
\big%
)=\mathbf{0}  \label{sys2}
\end{equation}%
that split to the two equations%
\begin{equation*}
\begin{array}{rl}
\mathbf{N}_{\mathbf{J}_{1},\mathbf{M}}^{T}%
\big%
(\mathbf{M}\left( \mathbf{q}\right) {\ddot{\mathbf{q}}}+\mathbf{C}\left( {%
\dot{\mathbf{q}},}\mathbf{q}\right) {\dot{\mathbf{q}}+{\mathbf{P}\left( {%
\mathbf{q}}\right) }+}\mathbf{Q}\left( {\dot{\mathbf{q}},\mathbf{q},t}%
\right) -\mathbf{u}\left( {\mathbf{q},t}\right) 
\big%
)=\mathbf{0}, & \text{ for }t<0 \\ 
&  \\ 
\mathbf{N}_{\mathbf{J}_{1},\mathbf{M}}^{T}%
\big%
(\mathbf{M}\left( \mathbf{q}\right) {\ddot{\mathbf{q}}}+\mathbf{C}\left( {%
\dot{\mathbf{q}},}\mathbf{q}\right) {\dot{\mathbf{q}}+{\mathbf{P}\left( {%
\mathbf{q}}\right) }+}\mathbf{Q}\left( {\dot{\mathbf{q}},\mathbf{q},t}%
\right) +\mathbf{J}_{2}^{T}\left( \mathbf{q}\right) \mathbf{\lambda }_{2}-%
\mathbf{u}\left( {\mathbf{q},t}\right) 
\big%
)=\mathbf{0}, & \text{ for }t\geq 0%
\end{array}%
\end{equation*}%
where $\mathbf{\lambda }_{2}$ is the part of $\mathbf{\lambda }$ of the
constraint reactions corresponding to the activated constraints, i.e. the
Jacobian $\mathbf{J}_{2}$. With $\mathbf{J}{\ddot{\mathbf{q}}=-\dot{\mathbf{J%
}}\dot{\mathbf{q}}}$ the equations (\ref{sys2}) can be written as%
\begin{equation}
\mathbf{M}\left( \mathbf{q}\right) {\ddot{\mathbf{q}}}+\mathbf{N}_{\mathbf{J}%
_{1},\mathbf{M}}^{T}\left( \mathbf{q}\right) 
\big%
(\mathbf{C}\left( {\dot{\mathbf{q}},}\mathbf{q}\right) {\dot{\mathbf{q}}}+{%
\mathbf{P}\left( {\mathbf{q}}\right) }+\mathbf{Q}\left( {\dot{\mathbf{q}},%
\mathbf{q},t)}+\mathbf{J}^{T}\left( \mathbf{q},t\right) \mathbf{\lambda }%
\right) 
\big%
){+\mathbf{J}_{1}^{T}\left( \mathbf{J}_{1}\mathbf{M}^{.-1}\mathbf{J}%
_{1}^{T}\right) ^{-1}{\dot{\mathbf{J}}}}_{1}{{\dot{\mathbf{q}}}}=\mathbf{N}_{%
\mathbf{J}_{1},\mathbf{M}}^{T}\mathbf{u}\left( {\mathbf{q},t}\right) .
\end{equation}%
These are actually the equations (\ref{eq1}) after the switching. The
momentum balance then reads%
\begin{equation}
\int_{0}^{\varepsilon }\mathbf{M}\left( \mathbf{q}\right) {\ddot{\mathbf{q}}%
dt}+\int_{0}^{\varepsilon }\mathbf{N}_{\mathbf{J}_{1},\mathbf{M}}^{T}\left( 
\mathbf{q}\right) 
\big%
(\mathbf{C}\left( {\dot{\mathbf{q}},}\mathbf{q}\right) {\dot{\mathbf{q}}}+{%
\mathbf{P}\left( {\mathbf{q}}\right) +\mathbf{Q}}\left( {\dot{\mathbf{q}},%
\mathbf{q},t}\right) 
\big%
)dt+\int_{0}^{\varepsilon }\mathbf{N}_{\mathbf{J}_{1},\mathbf{M}}^{T}\left( 
\mathbf{q}\right) \mathbf{J}_{2}^{T}\left( \mathbf{q}\right) \mathbf{\lambda 
}_{2}dt=\int_{0}^{\varepsilon }\mathbf{N}_{\mathbf{J}_{1},\mathbf{M}}^{T}%
\mathbf{u}\left( {\mathbf{q},t}\right) dt.  \label{sys3}
\end{equation}%
With the assumed smoothness of the persistent constraints (assumption \ref%
{assumption2}), $\mathbf{N}_{\mathbf{J}_{1},\mathbf{M}}$ is smooth, and
neglecting the integral of $\mathbf{Q}$ (assumption \ref{assumption1}), the
second integral term vanishes. Presuming smoothness of the activated
constraints for $t\geq 0$, then (\ref{sys3}) yields%
\begin{equation}
\mathbf{M}\left( \mathbf{q}_{0}\right) {\Delta \dot{\mathbf{q}}}+\mathbf{N}_{%
\mathbf{J}_{1},\mathbf{M}}^{T}\left( \mathbf{q}_{0}\right) \mathbf{J}%
_{2}^{T}\left( \mathbf{q}_{0}\right) \Lambda _{2}=\mathbf{N}_{\mathbf{J}_{1},%
\mathbf{M}}^{T}\left( \mathbf{q}_{0}\right) \mathbf{U}\left( \mathbf{q}%
_{0}\right)
\end{equation}%
with the impulsive constraint force $\Lambda _{2}:=\int_{0}^{\varepsilon }%
\mathbf{\lambda }_{2}dt$.

If $\mathbf{u}$ is smooth at $t_{0}$ (no discontinuous control forces), the
impulsive force $\mathbf{U}\left( \mathbf{q}_{0}\right) $ vanishes.

\subsubsection{Kinematic Compatibility}

The persistent constraints can be written as%
\begin{equation}
{{\mathbf{J}}_{1}\left( \mathbf{q}_{0}\right) \dot{\mathbf{q}}}={\mathbf{J}}%
_{1}\left( \mathbf{q}_{0}\right) {\dot{\mathbf{q}}}_{+}-{{\mathbf{J}}%
_{1}\left( \mathbf{q}_{0}\right) \Delta \dot{\mathbf{q}}}=\mathbf{0},
\label{cons1}
\end{equation}%
and the additional constraints for $t\geq 0$ as%
\begin{equation}
{{\mathbf{J}}_{2}\left( \mathbf{q}_{0}\right) \dot{\mathbf{q}}}_{+}={\mathbf{%
J}}_{2}\left( \mathbf{q}_{0}\right) {\Delta \dot{\mathbf{q}}+{\mathbf{J}}%
_{2}\left( \mathbf{q}_{0}\right) \dot{\mathbf{q}}}_{-}=\mathbf{0}.
\label{cons2}
\end{equation}%
The constraints (\ref{cons1}) must be satisfied throughout the switching. A
particular solution of (\ref{cons1}) is ${\dot{\mathbf{q}}}_{+}=\mathbf{J}%
_{1,\mathbf{M}}^{+}\left( \mathbf{q}_{0}\right) {\mathbf{J}}_{1}\left( 
\mathbf{q}_{0}\right) {\Delta \dot{\mathbf{q}}}$. Inserting this into (\ref%
{cons2}) yields 
\begin{equation}
{{\mathbf{J}}_{2}\left( \mathbf{q}_{0}\right) }\mathbf{N}_{\mathbf{J}_{1},%
\mathbf{M}}{\left( \mathbf{q}_{0}\right) \Delta \dot{\mathbf{q}}=-{\mathbf{J}%
}_{2}\left( \mathbf{q}_{0}\right) \dot{\mathbf{q}}}_{-}.
\end{equation}

\subsubsection{Overall Transition Condition}

The momentum balance together with the kinematic compatibility conditions
can be summarized as%
\begin{equation}
\left( 
\begin{array}{cc}
\mathbf{M}\left( \mathbf{q}_{0}\right) & \mathbf{N}_{\mathbf{J}_{1},\mathbf{M%
}}^{T}\left( \mathbf{q}_{0}\right) \mathbf{J}_{2}^{T}\left( \mathbf{q}%
_{0}\right) \\ 
{{\mathbf{J}}_{2}\left( \mathbf{q}_{0}\right) }\mathbf{N}_{\mathbf{J}_{1},%
\mathbf{M}}{\left( \mathbf{q}_{0}\right) } & \mathbf{0}%
\end{array}%
\right) \left( 
\begin{array}{c}
{\Delta \dot{\mathbf{q}}} \\ 
\Lambda _{2}%
\end{array}%
\right) =\left( 
\begin{array}{c}
\mathbf{U}\left( \mathbf{q}_{0}\right) \\ 
-{\mathbf{J}}_{2}\left( \mathbf{q}_{0}\right) {\dot{\mathbf{q}}}_{-}%
\end{array}%
\right) .  \label{cond1}
\end{equation}%
The simple case of an MBS subjected to a single set of constraints, which is
activated at $t=0$, is included with $\mathbf{N}_{\mathbf{J}_{1},\mathbf{M}}=%
\mathbf{I}$. The so determined ${\Delta \dot{\mathbf{q}}}$ is the admissible
velocity jump such that the momentum balance is satisfied. The impulsive
constraint force $\Lambda _{2}$ represents the change in the system's
generalized momentum. As a by-product, the impulsive constraint force $%
\Lambda _{2}$ caused by the event is determined.

\begin{itemize}
\item The compatibility condition (\ref{cond1}) yields the full (redundant)
state of the system. It can be used in conjunction with any dynamics
formulation.

\item The square coefficient matrix has dimension $n+m_{2}$ since only the $%
m_{2}$ (projected) constraints are included that are added at the event.
Solving (\ref{cond1}) requires inversion of the $\left( n+m_{2}\right)
\times \left( n+m_{2}\right) $ matrix. The determination of the null-space
projector requires inversion of a $m_{1}\times m_{1}$ matrix. This inverse
exists with assumption \ref{assumption2}.

\item The system (\ref{cond1}) is solved at the switching event. If the EOM
are evaluated with an $O\left( n\right) $ method, then possibly also the
system (\ref{cond1}) can be solved using this method. The apparent obstacle
is the null-space projector. Evaluating the null-space projector requires
computing the pseudoinverse $\mathbf{J}_{\mathbf{M}}^{+}=\mathbf{M}^{-1}%
\mathbf{J}^{T}\left( \mathbf{JM}^{-1}\mathbf{J}^{T}\right) ^{-1}$, and thus
the inverse of the mass matrix. The complexity depends on the specific form
of the constraints. These are very simple for instance, if joints are locked
(see sections 6 and 7).

\item If the constraints $h_{1}\left( \mathbf{q}\right) =\mathbf{0}$ and $%
h_{2}\left( \mathbf{q}\right) =\mathbf{0}$ are independent, i.e. $\ker 
\mathbf{J}_{1}\cap \mathrm{range}~\mathbf{J}_{2}=\emptyset $ (assumption \ref%
{assumption3}), and $\mathbf{J}_{2}$ has full rank (assumption \ref%
{assumption2}), then the coefficient matrix in (\ref{cond1}) has full rank $%
n+m_{2}$.
\end{itemize}

\subsection{Transition Condition in Terms of Minimal Coordinates}

\subsubsection{Momentum Balance}

The above condition (\ref{cond1}) involves the null-space projector, and
thus requires the determination of $\mathbf{M}^{-1}$. Even if this is only
necessary at the switching points, for large systems this may be infeasible
to be performed. An alternative formulation can be derived in terms of
minimal coordinates resorting to the concept of Voronets equations (\ref%
{Voronets}). The latter cannot be applied unaltered to quasi-scleronomic
constraints, however. For the special case of activation of additional
constraints, the persistent constraints $h_{1}\left( \mathbf{q}\right) =%
\mathbf{0}$ can be used to formulate EOM on $V_{-}$. The additional
constraints $h_{2}\left( \mathbf{q}\right) =\mathbf{0}$ then restrict the
EOM to $V_{+}\subset V_{-}$.

Starting from the Lagrange equations (\ref{Lagrange2}), the Voronets
equations on $V_{-}$ are%
\begin{equation}
\overline{\mathbf{M}}\left( \mathbf{q}\right) {\ddot{\mathbf{s}}}+\overline{%
\mathbf{C}}\left( {\dot{\mathbf{q}},}\mathbf{q}\right) {\dot{\mathbf{s}}+%
\overline{\mathbf{Q}}}\left( {\dot{\mathbf{q}},\mathbf{q},t}\right) +\mathbf{%
F}_{1}^{T}\left( \mathbf{q}\right) \mathbf{J}^{T}\left( \mathbf{q},t\right) 
\mathbf{\lambda }=\overline{\mathbf{u}}\left( {\mathbf{q},t}\right)
\label{VoronetsProj}
\end{equation}%
that split into%
\vspace{-3ex}%
\begin{equation*}
\begin{array}{rl}
\overline{\mathbf{M}}\left( \mathbf{q}\right) {\ddot{\mathbf{s}}}+\overline{%
\mathbf{C}}\left( {\dot{\mathbf{q}},}\mathbf{q}\right) {\dot{\mathbf{s}}+%
\overline{\mathbf{Q}}}\left( {\dot{\mathbf{q}},\mathbf{q},t}\right) =%
\overline{\mathbf{u}}\left( {\mathbf{q},t}\right) , & \text{ for }t<0 \\ 
&  \\ 
\overline{\mathbf{M}}\left( \mathbf{q}\right) {\ddot{\mathbf{s}}}+\overline{%
\mathbf{C}}\left( {\dot{\mathbf{q}},}\mathbf{q}\right) {\dot{\mathbf{s}}+%
\overline{\mathbf{Q}}}\left( {\dot{\mathbf{q}},\mathbf{q},t}\right) +\mathbf{%
F}_{1}^{T}\left( \mathbf{q}\right) \mathbf{J}_{2}^{T}\left( \mathbf{q}%
\right) \mathbf{\lambda }_{2}=\overline{\mathbf{u}}\left( {\mathbf{q},t}%
\right) , & \text{ for }t\geq 0%
\end{array}%
\end{equation*}%
with $\mathbf{F}_{1}$ being the orthogonal complement to $\mathbf{J}_{1}$,
and $\mathbf{\lambda }_{2}$ are the Lagrange multipliers accounting for the
additional constraints not captured by the projection to $V_{-}$. This
resembles the formulation in redundant coordinates (\ref{sys2}), but now the
EOM already projected to $V_{-}$ are subjected to additional constraints,
whereas in (\ref{sys2}) the projector changes. The equations (\ref%
{VoronetsProj}) may be considered as constrained Voronets equations. The
crucial difference is that here locally valid set of independent coordinates 
$\mathbf{s}$ must be selected explicitly.

Proceeding as above, the momentum balance during the switching leads to%
\begin{equation}
\int_{0}^{\varepsilon }\overline{\mathbf{M}}\left( \mathbf{q}\right) {\ddot{%
\mathbf{s}}dt}+\int_{0}^{\varepsilon }%
\big%
(\overline{\mathbf{C}}\left( {\dot{\mathbf{q}},}\mathbf{q}\right) {\dot{%
\mathbf{q}}+\overline{\mathbf{Q}}}\left( {\dot{\mathbf{q}},\mathbf{q},t}%
\right) 
\big%
)dt+\int_{0}^{\varepsilon }\mathbf{F}_{1}^{T}\left( \mathbf{q}\right) 
\mathbf{J}_{2}^{T}\left( \mathbf{q}\right) \mathbf{\lambda }%
_{2}dt=\int_{0}^{\varepsilon }\overline{\mathbf{u}}\left( {\mathbf{q},t}%
\right) dt.
\end{equation}%
It is presumed that $\mathbf{s}$ are valid local coordinates, so that $%
\mathbf{F}$ is smooth, that $\mathbf{J}_{1}$ is full rank so that $\mathbf{C}%
{\dot{\mathbf{q}}}$ and $\mathbf{Q}$ are smooth. The momentum balance then
yields%
\begin{equation}
\overline{\mathbf{M}}\left( \mathbf{q}_{0}\right) {\Delta \dot{\mathbf{s}}}+%
\mathbf{F}_{1}^{T}\left( \mathbf{q}_{0}\right) \mathbf{J}_{2}^{T}\left( 
\mathbf{q}_{0}\right) \Lambda _{2}dt=\overline{\mathbf{U}}\left( \mathbf{q}%
_{0}\right)  \label{MomentumBal2}
\end{equation}%
where ${\Delta \dot{\mathbf{s}}}={\dot{\mathbf{s}}}_{+}-{\dot{\mathbf{s}}}%
_{-}$ is the change in the independent generalized velocities, $\Lambda
_{2}:=\int_{0}^{\varepsilon }\mathbf{\lambda }_{2}dt$ is the impulsive
constraint force collocated with the minimal coordinates, and $\overline{%
\mathbf{U}}\left( \mathbf{q}_{0}\right) =\int_{0}^{\varepsilon }\overline{%
\mathbf{u}}\left( {\mathbf{q},t}\right) dt=\mathbf{F}_{1}^{T}\left( \mathbf{q%
}_{0}\right) \int_{0}^{\varepsilon }\mathbf{u}\left( {\mathbf{q},t}\right)
dt=\mathbf{F}_{1}^{T}\left( \mathbf{q}_{0}\right) \mathbf{U}\left( \mathbf{q}%
_{0}\right) $. The increment of the generalized velocities is thus ${\Delta 
\dot{\mathbf{q}}}=\mathbf{F}_{1}\left( \mathbf{q}_{0}\right) {\Delta \dot{%
\mathbf{s}}}$.

\subsubsection{Kinematic Compatibility}

The persistent velocity constraints are already incorporated in the Voronets
equations on $V_{-}$. It remains to account for the additionally activated
constraints. After the switching event these are 
\begin{eqnarray}
\mathbf{0} &=&{{\mathbf{J}}_{2}\left( \mathbf{q}_{0}\right) \dot{\mathbf{q}}}%
_{+}={{\mathbf{J}}_{2}\left( \mathbf{q}_{0}\right) \mathbf{F}}_{1}(\mathbf{q}%
_{0}){\dot{\mathbf{s}}}_{+}  \notag \\
&=&{\mathbf{J}}_{2}\left( \mathbf{q}_{0}\right) {\mathbf{F}}_{1}(\mathbf{q}%
_{0})\Delta {\dot{\mathbf{s}}}+{{\mathbf{J}}_{2}\left( \mathbf{q}_{0}\right) 
{\mathbf{F}}_{1}(\mathbf{q}_{0}}){\dot{\mathbf{s}}}_{-}.  \label{kinconstr}
\end{eqnarray}%
The last equation relates the independent velocities before the event, ${%
\dot{\mathbf{s}}}_{-}$, to the change due to the switching, ${\Delta \dot{%
\mathbf{s}}}$.

\subsubsection{Overall Transition Condition}

Combining momentum balance (\ref{MomentumBal2}) with the kinematic
constraints (\ref{kinconstr}) gives rise to the overall condition

\begin{equation}
\left( 
\begin{array}{cc}
\overline{\mathbf{M}}\left( \mathbf{q}_{0}\right) & {\mathbf{F}}_{1}^{T}(%
\mathbf{q}_{0}){{\mathbf{J}}_{2}^{T}\left( \mathbf{q}_{0}\right) } \\ 
{{\mathbf{J}}_{2}\left( \mathbf{q}_{0}\right) \mathbf{F}}_{1}(\mathbf{q}_{0})
& \mathbf{0}%
\end{array}%
\right) \left( 
\begin{array}{c}
{\Delta \dot{\mathbf{s}}} \\ 
\Lambda _{2}%
\end{array}%
\right) =\left( 
\begin{array}{c}
\mathbf{U}\left( \mathbf{q}_{0}\right) \\ 
-{\mathbf{J}}_{2}\left( \mathbf{q}_{0}\right) {\mathbf{F}}_{1}(\mathbf{q}%
_{0}){\dot{\mathbf{s}}}_{-}%
\end{array}%
\right) .  \label{cond2}
\end{equation}%
This is the compatibility condition on the generalized momentum and the
generalized velocity. A solution ${\Delta \dot{\mathbf{s}}}$ of (\ref{cond2}%
) satisfies the kinematic constraints while ensuring momentum balance.

This is a system of $n-m_{1}+m_{2}$ equations for the change of $n-m_{1}$
generalized velocities ${\Delta \dot{\mathbf{s}}}$ and the $m_{2}$ impulsive
constraint forces $\Lambda _{2}$. The system (\ref{cond2}) has an obvious
interpretation as momentum balance of a constrained MBS with local DOF $%
n-m_{1}$ that is subjected to $m_{2}$ additional constraints.

\begin{itemize}
\item Solving (\ref{cond2}) requires inverting the $\left(
n-m_{1}+m_{2}\right) \times \left( n-m_{1}+m_{2}\right) $ coefficient matrix.

\item The computation of the orthogonal complement $\mathbf{F}_{1}$ requires
inversion of a $m_{1}\times m_{1}$ matrix.
\end{itemize}

\section{Example 1: Joint Locking in a Planar 3-Bar Pendulum%
\label{secExample1}%
}

\begin{figure}[h]
\begin{center}
\includegraphics[width=5cm]{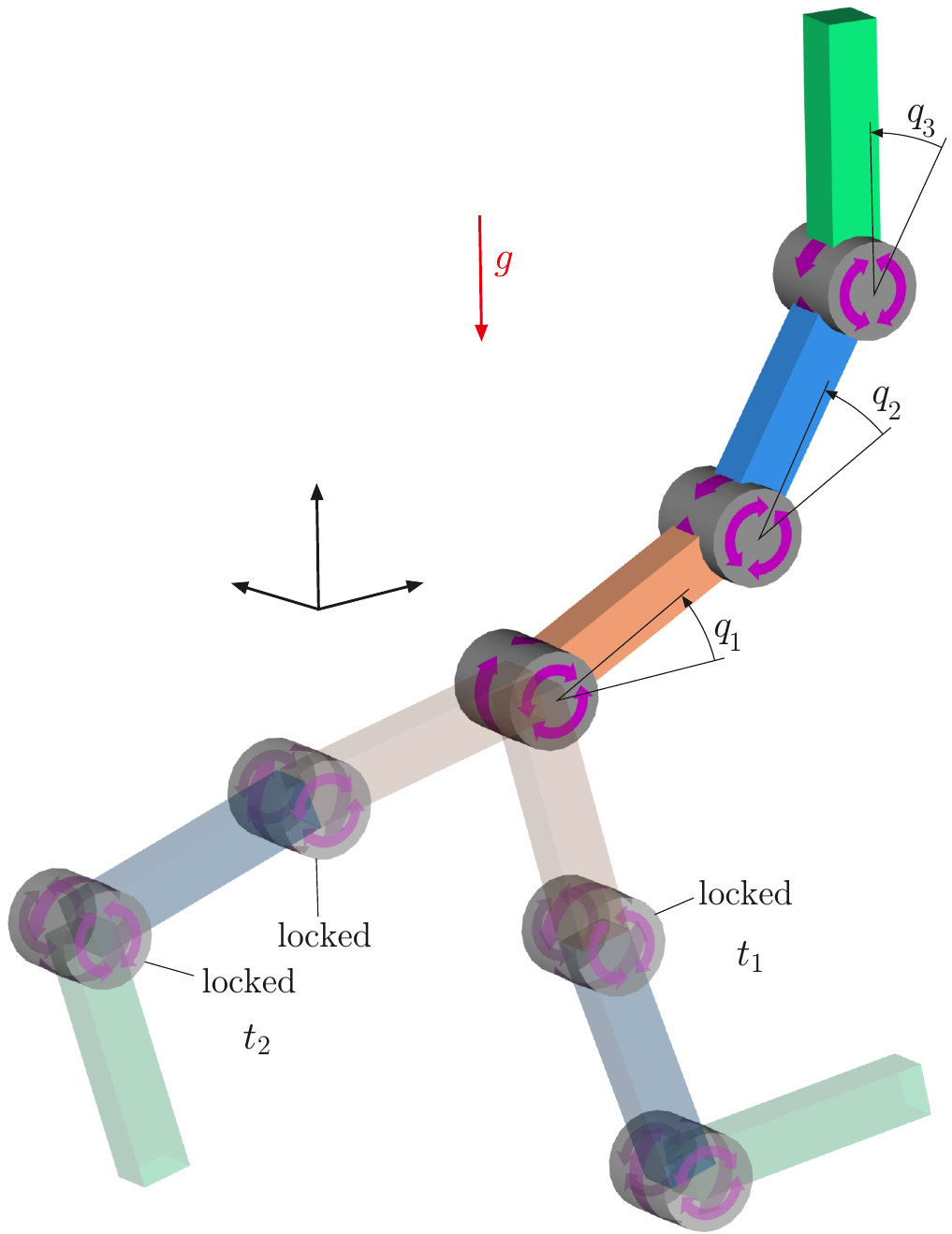}
\end{center}
\par
\vspace{-3ex}
\caption{Planar 3R-pendulum in its reference configurations $%
q_{1}=q_{2}=q_{3}=\protect\pi /6$. Also shown are the configurations at the
switching time $t_{1}=0.8$~s where joint 2 is locked, and at the switching
time $t_{2}=1.3~$s where joint 3 is locked.}
\label{fig3R}
\end{figure}

The first example is a planar 3R pendulum with parallel revolute joint axes
linked to the ground and subjected to gravity in fig. \ref{fig3R}. The solid
aluminium links have lengths 1~m and rectangular cross section of $0.2\times
0.2$~m. Accordingly, the mass of each link is $m_{i}=108$ kg, and the
principle inertia moments w.r.t. to the COM are $\Theta _{11}=0.72$ kg~m$%
^{2},\Theta _{22}=\Theta _{33}=9.36$ kg~m$^{2}$. The generalized coordinates
are the three joint angles. For the experiment, the system is released from
rest in the configuration $\mathbf{q}=\left( \pi /6,\pi /6,\pi /6\right) $
and set in motion by the gravity force ($g=9.81~$m/s$^{2}$). 

At $t_{1}=0.8~$s, the second joint is locked. The third joint is locked at $%
t_{2}=1.3$~s. This successive joint locking is described by the
quasi-scleronomic constraints 
\begin{equation}
h\left( \mathbf{q},t\right) =\left\{ 
\begin{array}{cl}
0, & t<t_{1} \\ 
& 
\vspace{-1.6ex}
\\ 
h_{1}\left( \mathbf{q}\right) , & t_{1}\leq t<t_{2} \\ 
& 
\vspace{-1.6ex}
\\ 
\left( 
\begin{array}{c}
h_{1}\left( \mathbf{q}\right) \\ 
h_{2}\left( \mathbf{q}\right)%
\end{array}%
\right) , & t_{2}\leq t%
\end{array}%
\right. ,\ \ \ \mathbf{J}\left( \mathbf{q},t\right) =\left\{ 
\begin{array}{cl}
0, & t<t_{1} \\ 
& 
\vspace{-1.6ex}
\\ 
\mathbf{J}_{1}, & t_{1}\leq t<t_{2} \\ 
& 
\vspace{-1.6ex}
\\ 
\left( 
\begin{array}{c}
\mathbf{J}_{1} \\ 
\mathbf{J}_{2}%
\end{array}%
\right) , & t_{2}\leq t\ \ .%
\end{array}%
\right.  \label{constrExample1}
\end{equation}%
with $h_{1}\left( \mathbf{q}\right) =q_{2}-q_{2}\left( t_{1}\right) $, and $%
h_{2}\left( \mathbf{q}\right) =q_{3}-q_{3}\left( t_{2}\right) $, and the
corresponding Jacobians%
\begin{equation}
\mathbf{J}_{1}=\left( 
\begin{array}{ccc}
0 & 1 & 0%
\end{array}%
\right) ,\ \ \mathbf{J}_{2}=\left( 
\begin{array}{ccc}
0 & 0 & 1%
\end{array}%
\right) .  \label{J}
\end{equation}

\paragraph{Minimal Coordinates}

Prior to $t_{1}$ the system is unconstrained, so that the transition
condition (\ref{sys1}) applies with ${\mathbf{J}}_{+}={\mathbf{J}}_{1}$.
Additionally, joint 3 is locked at $t_{2}$. Now the compatibility condition (%
\ref{cond2}) is invoked with$\ \mathbf{J}_{2}$ in (\ref{J}), and with the
orthogonal complement $\mathbf{F}_{1}=\left( 
\begin{array}{cc}
1 & 0 \\ 
0 & 0 \\ 
0 & 1%
\end{array}%
\right) $ to $\mathbf{J}_{1}$. The minimal coordinates are $\mathbf{s}%
=\left( q_{1},q_{1}\right) $. The latter are the remaining independent joint
variables after the locking of joint 2 at $t_{1}$. Notice that, when joints
in an open kinematic chain are locked, the set of minimal coordinates in the
transition equations is unique.

\paragraph{Redundant Coordinates}

When (\ref{cond1}) is used as transition condition for the additional joint
locking at $t_{2}$, the null-space projector to $\mathbf{J}_{1}$, and thus
the pseudoinverse is required. From the special form of $\mathbf{J}_{1}$
follows $\mathbf{J}_{1\mathbf{M}}^{+}=\mathbf{M}^{-1}\mathbf{J}%
_{1}^{T}\left( \mathbf{J}_{1}\mathbf{M}^{.-1}\mathbf{J}_{1}^{T}\right)
^{-1}=\left( \bar{m}_{12}/\bar{m}_{22},1,\bar{m}_{32}/\bar{m}_{22}\right)
^{T}$, where $\bar{m}_{ij}$ are the entries of the inverse $\mathbf{M}^{-1}$%
, and thus%
\begin{equation*}
\mathbf{N}_{\mathbf{J}_{1},\mathbf{M}}=\left( 
\begin{array}{ccc}
1 & -\bar{m}_{12}/\bar{m}_{22} & 0 \\ 
0 & 0 & 0 \\ 
0 & -\bar{m}_{32}/\bar{m}_{22} & 1%
\end{array}%
\right)
\end{equation*}

\paragraph{Results}

\begin{figure}[tbh]
\begin{center}
{\small a)}\hspace{-1.5ex}%
\includegraphics[width=7.5cm,height=5cm]{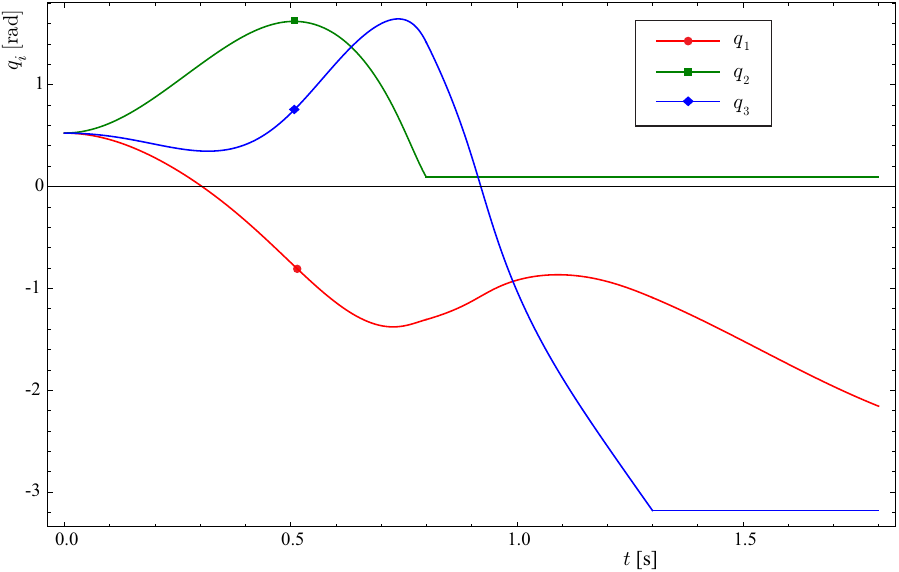}%
~~~~{\small b)}\hspace{-2.5ex}%
\includegraphics[width=7.5cm,height=5cm]{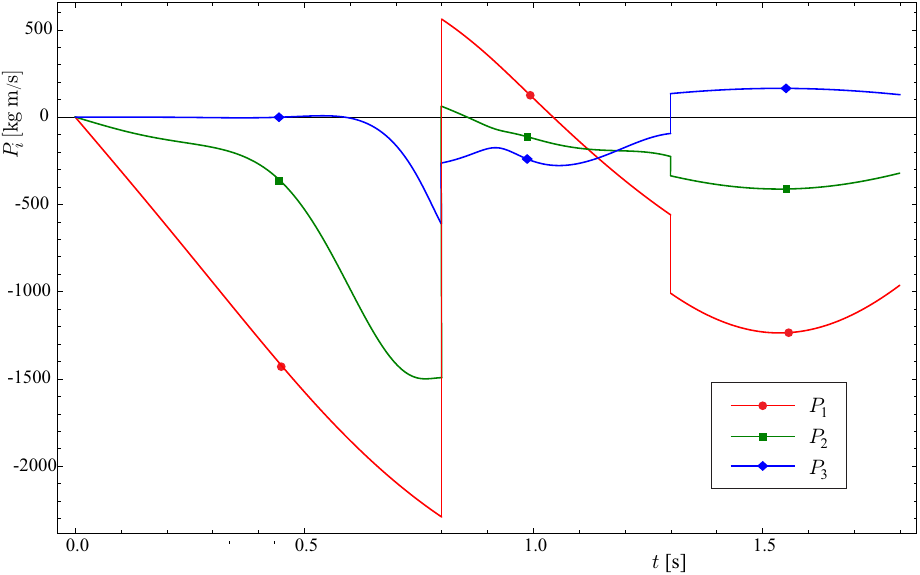}
\end{center}
\par
\vspace{-5ex}
\caption{a) Joint angles, and b) generalized momenta of the planar
3R-pendulum when integrated without momentum consistent switching.}
\label{InconsistentMomentumFig}
\end{figure}

\begin{figure}[tbh]
\begin{center}
{\small a)}\hspace{-1.5ex}%
\includegraphics[width=7.5cm,height=5cm]{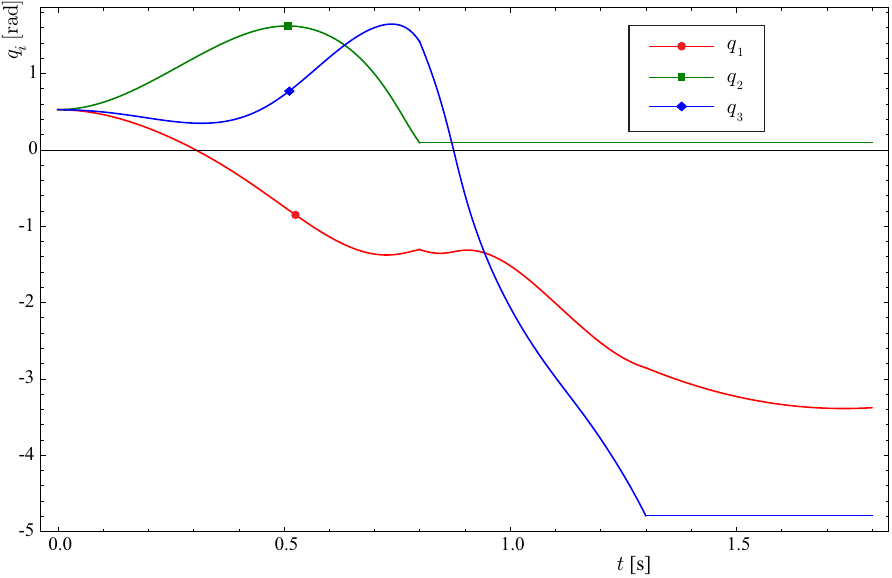}~~~~{\small b)}%
\hspace{-2.5ex}%
\includegraphics[width=7.5cm,height=5cm]{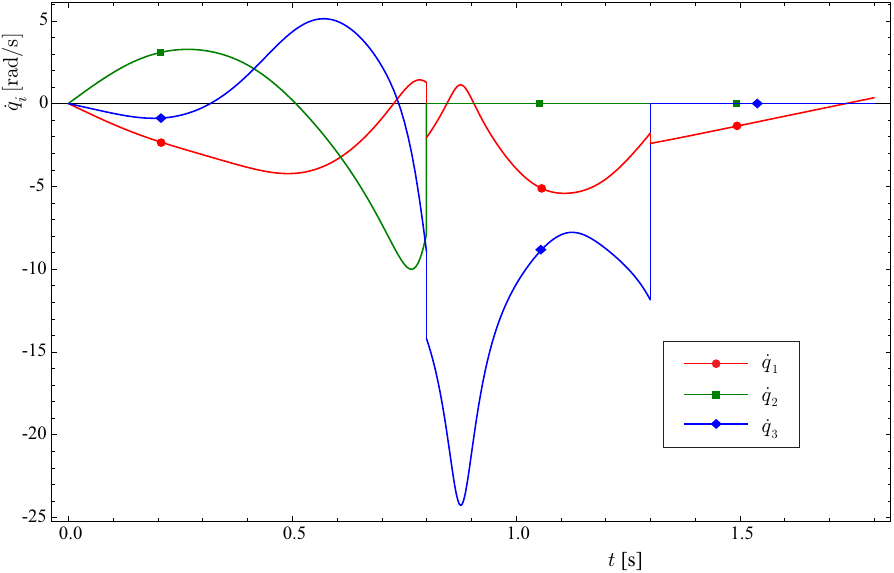}
\end{center}
\par
\vspace{-3ex}
\caption{a) Joint angles, and b) joint rates of the planar 3R-pendulum.}
\label{qFig}
\end{figure}

\begin{figure}[h]
\begin{center}
{\small a)}\hspace{-1.5ex}%
\includegraphics[width=7.5cm,height=5cm]{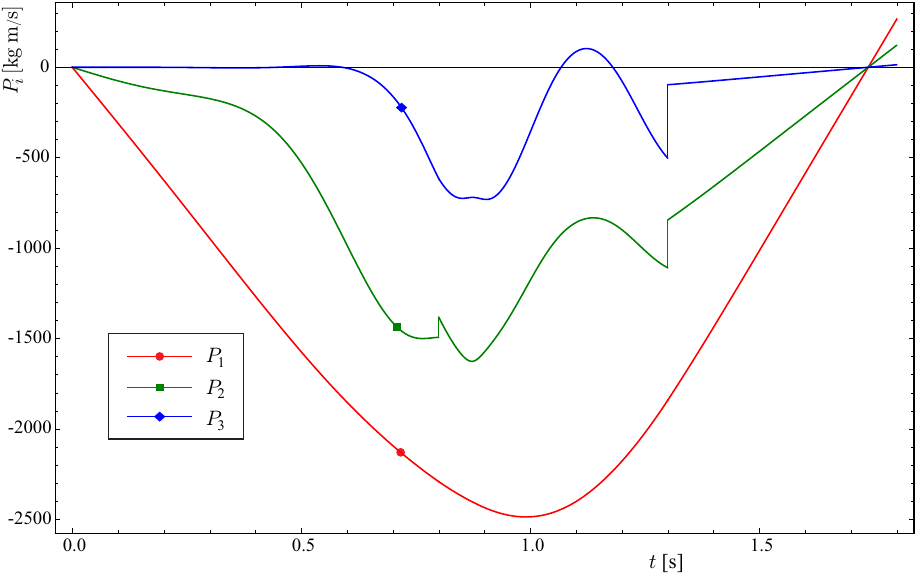}~~~%
{\small b)}\hspace{-1.1ex}%
\includegraphics[width=7.4cm,height=5cm]{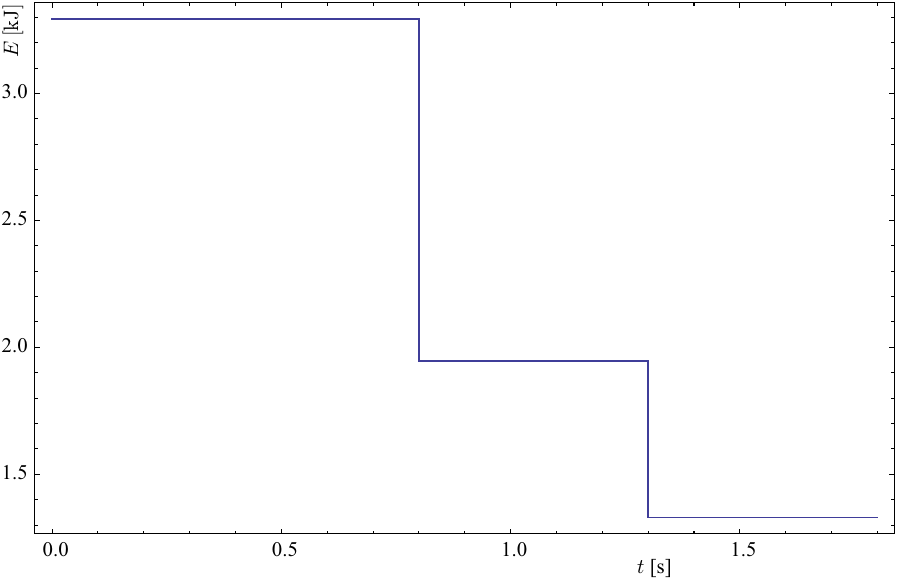}
\end{center}
\par
\vspace{-4ex}
\caption{a) Generalized momenta, and b) total energy of the planar
3R-pendulum.}
\label{MomentumFig}
\end{figure}

The motion equations are integrated with an explicit Runge-Kutta 4
integration scheme using a fixed time step size of $10^{-4}$ s. The
transition conditions are invoked at the switching events at $t_{1}$ and $%
t_{2}$. In order to show the need for a momentum consistent switching, the
EOM are integrated where the constraints (\ref{constrExample1}) are simply
imposed and $\dot{q}_{2}$ and $\dot{q}_{3}$ is set to zero at $t_{1}$ and $%
t_{2}$, respectively. The resulting time evolution of the joint angles and
the generalized momenta is shown in fig. \ref{InconsistentMomentumFig}.
Clearly visible is that the momenta of the unlocked joints is not conserved.
Next the EOM are integrated using the consistent switching. To this end,
both formulations (\ref{cond1}) and (\ref{cond2}) were evaluated, and lead
to identical results. The time evolution of the generalized coordinates is
shown in fig. \ref{qFig}a), and the corresponding generalized velocities in
fig. \ref{qFig}b). The velocity jumps are such that the momenta of the
unlocked joints are continuous. Fig. \ref{MomentumFig}a) shows the
components of the generalized momentum vector $\mathbf{P}=\mathbf{F}_{i}^{T}%
\mathbf{M}\left( \mathbf{q}\right) {\dot{\mathbf{q}}}$. These are merely of
the unlocked joints. At the switching points, the total energy jumps whereas
it is preserved otherwise (fig. \ref{MomentumFig}b)). Notice that the
geometric locking constraint is exactly satisfied because the transition
conditions (\ref{cond1}) yields $\dot{q}_{2}=0$, which is exactly integrated
to $q_{2}=q_{2}\left( t_{1}\right) $, for $t\geq t_{1}$, and analogously for 
$q^{3}$. Comparing fig. \ref{InconsistentMomentumFig}a) and fig. \ref{qFig}%
a) the difference of the obtained joint trajectories is apparent. 

\newpage%

\section{Example 2: Joint Locking in a Planar 3-Bar Pendulum%
\label{secExample2}%
}

\begin{figure}[h]
\begin{center}
\hspace{-1.5ex}\includegraphics[width=5cm]{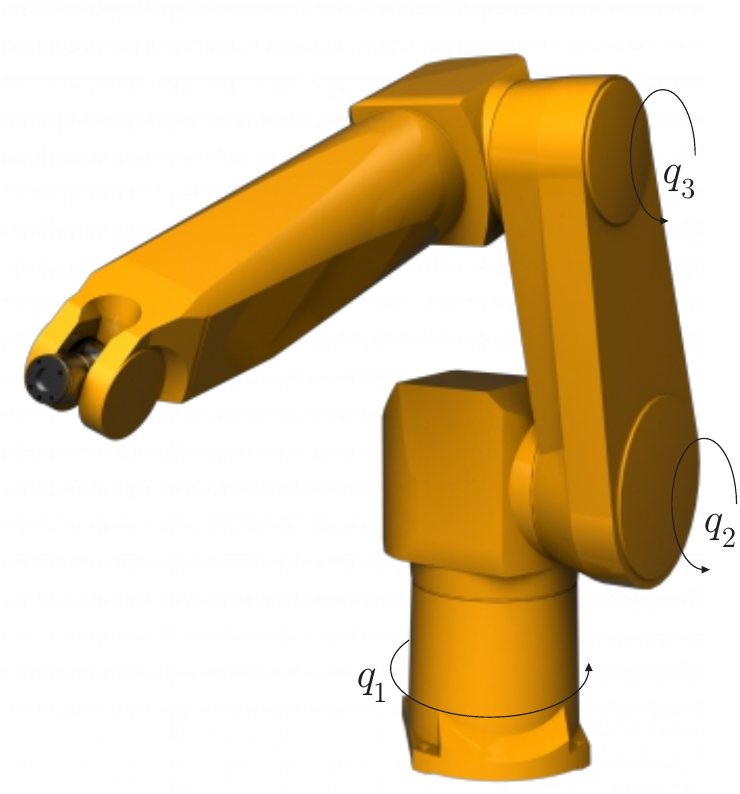}
\end{center}
\par
\vspace{-4ex}
\caption{RX 130L robotic manipulator in the reference configuration $%
q_1=q_2=q_3=0$.}
\label{RX130RefFig}
\end{figure}

The second example is an industrial 6 DOF serial robot St\"{a}ubli RX130L
(fig. \ref{RX130RefFig}). All geometric and inertia parameter were
identified and used in the dynamic model. The serial robot can be split into
a wrist (the last three revolute joints) and the serial chain, consisting of
the first three revolute joints, which achieves the positioning of the
wrist. Since wrist locking has no significant effect on the overall motion,
a simplified scenario is considered where joint 1 is locked at $t=0.05$ s,
then joint 2 is locked at $t=0.1$ s, and finally joint 3 is locked at $t=0.15
$ s. In the following figures the motor angles and rates are shown that are
related to the joint angles and rates by a gear ration of 100. 

\begin{figure}[h]
\begin{center}
{\small a)}\hspace{-1.5ex}%
\includegraphics[width=7.5cm]{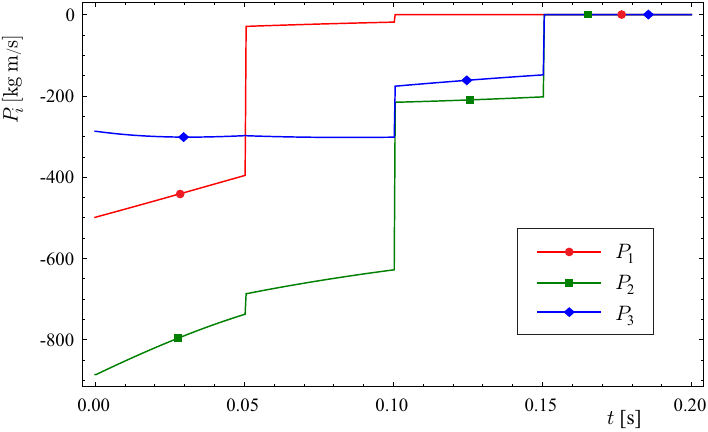}~~~%
{\small b)}\hspace{-1.1ex}%
\includegraphics[width=7.4cm]{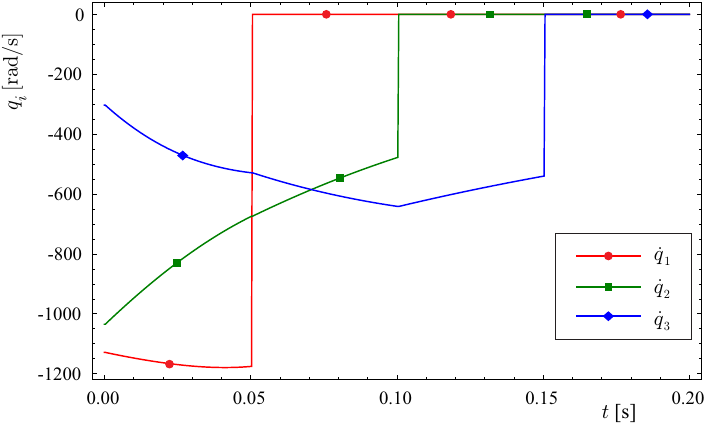}
\end{center}
\par
\vspace{-4ex}
\caption{a) Generalized momenta, and b) motor rates $\dot{q}_{1},\dot{q}_{2},%
\dot{q}_{3}$ for the RX130L manipulator when no momentum inconsistent
transition is performed.}
\label{RX130InconsistentFig}
\end{figure}

\begin{figure}[h]
\begin{center}
{\small a)}\hspace{-1.5ex}%
\includegraphics[width=7.5cm]{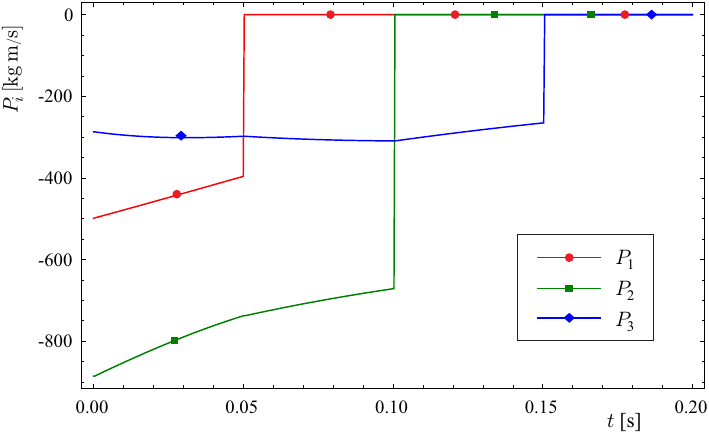}~~~{\small b)}%
\hspace{-1.1ex}\includegraphics[width=7.4cm]{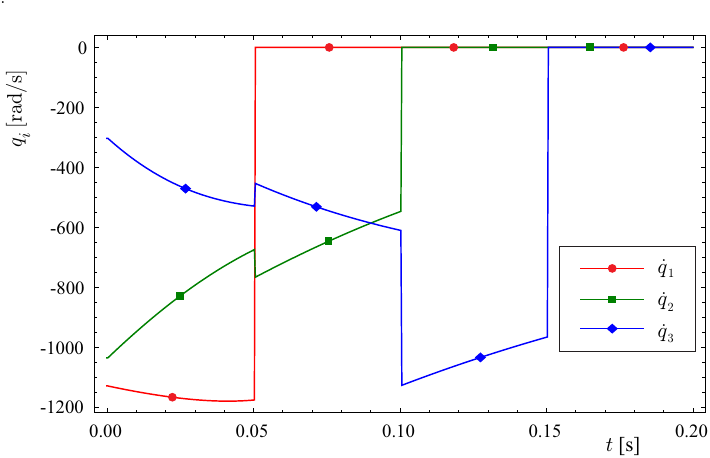}
\end{center}
\par
\vspace{-4ex}
\caption{a) Generalized momenta, and b) motor rates $\dot{q}_{1},\dot{q}_{2},%
\dot{q}_{3}$ for the RX130L manipulator when using momentum consistenten
update.}
\label{RX130Fig}
\end{figure}

\begin{figure}[h]
\begin{center}
{\small a)}\hspace{-1.5ex}%
\includegraphics[width=7.5cm]{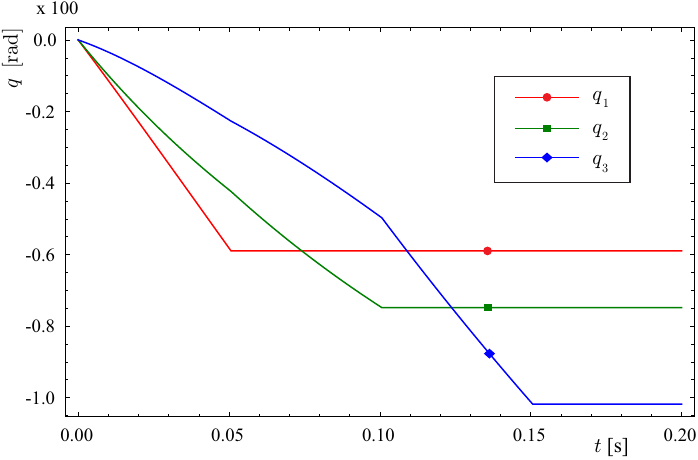}~~~{\small b)}\hspace{%
-1.1ex}\includegraphics[width=7.4cm]{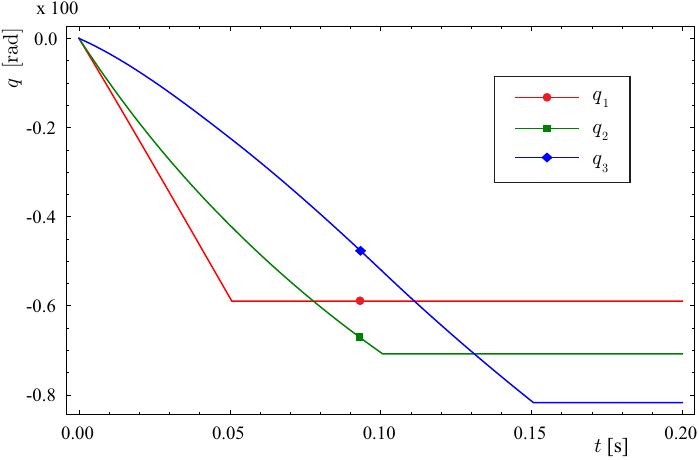}
\end{center}
\par
\vspace{-4ex}
\caption{Motor joint angles ${q}_{1},{q}_{2},{q}_{3}$ for the RX130L
manipulator when a) momentum consistency is enusured, and b) when no
momentum inconsistent transition is performed.}
\label{RX130qFig}
\end{figure}

Fig. \ref{RX130InconsistentFig} shows the generalized velocities and the
generalized momenta of the unlocked motors when no momentum consistent
update is used. Using the momentum consistent update scheme yields the motor
velocities and generalized momenta in fig. \ref{RX130Fig}. The respective
motor angles as shown in fig. \ref{RX130qFig}. From the latter it is
apparent that the terminal position at which the robot comes to a halt is
different. This is indeed a crucial safety aspect as this decides about
possible impact with environment and human operator. Moreover, the dynamics
simulation shall provide a means for safe motion planning ensuring that at
any time the system can be stopped when a potential collision is detected. 

\newpage

\section{Conclusion}

The time integration of the motion equations governing the dynamics of
variable topology mechanisms (VTM) must ensure the compatibility of the
generalized momentum and velocity at the switching events. In this paper a
compatibility condition for the general case of successive activation of
multiple constraints is presented. The condition is formulated in terms of
the full set of (due to the constraint activation) redundant coordinates as
well as in terms of minimal coordinates. The formulations have different
computational complexity. Both formulations can be evaluated independently
from the actual time integration method. The method can be easily extended
to account for non-holonomic velocities $\mathbf{v}$ related to generalized
speed by $\mathbf{v}=\mathbf{A}\left( \mathbf{q}\right) {\dot{\mathbf{q}}}$.
In this paper it is assumed that the constraints are non-redundant. As
future work, this condition could be relaxed and a corresponding formulation
derived.

As example, the locking of joints of a planar 3R mechanism and an industrial
6R serial manipulators are reported. The method has been applied to various
other systems, including a metamorphic robotic hand whose palm consists of a
spherical 5-bar linkage imitating the inherent mobility of the human hand.
Other possible applications include docking maneuvers.

The formulation applies also to MBS comprising flexible bodies, in which
case the generalized coordinates include modal/nodal coordinates. Then, as
always when constraining flexible MBS, the constraints may lead to
artificial mode locking.

\section*{Acknowledgement}

The author acknowledges that this work has been partially supported by the
Austrian COMET-K2 program of the Linz Center of Mechatronics (LCM).

\section*{Nomenclature}

\begin{tabular}{|lll|}
\hline
$n$ & -$%
\hspace{-1.6ex}%
$ & number joint variables \\ 
$m$ & -$%
\hspace{-1.6ex}%
$ & number of constraints \\ 
$\mathbf{q}\in {\mathbb{V}}^{n}$ & -$%
\hspace{-1.6ex}%
$ & vector of joint variables \\ 
${\dot{\mathbf{q}}\in {\mathbb{R}}^{n}}$ & -$%
\hspace{-1.6ex}%
$ & generalized velocity \\ 
${\Delta \dot{\mathbf{q}}}$ & -$%
\hspace{-1.6ex}%
$ & jump of generalized velocity due to constraint switching \\ 
$\delta _{\text{loc}}\left( \mathbf{q}\right) 
\hspace{-1.8ex}%
$ & -$%
\hspace{-1.6ex}%
$ & finite local DOF at $\mathbf{q}$ \\ 
$\delta _{\text{diff}}\left( \mathbf{q}\right) 
\hspace{-1.8ex}%
$ & -$%
\hspace{-1.6ex}%
$ & differential DOF at $\mathbf{q}$ \\ 
$V\subset {\mathbb{V}}^{n}$ & -$%
\hspace{-1.6ex}%
$ & Configuration space (c-space) of a MBS defined by the constraints \\ 
$h\left( \mathbf{q},t\right) $ & -$%
\hspace{-1.6ex}%
$ & quasi-scleronomic constraint function \\ 
$h_{-}$ & -$%
\hspace{-1.6ex}%
$ & constraint function before the event \\ 
$h_{+}$ & -$%
\hspace{-1.6ex}%
$ & constraint function after the event \\ 
$\mathbf{J}$ & -$%
\hspace{-1.6ex}%
$ & constraint Jacobian \\ 
$\mathbf{J}_{-}$ & -$%
\hspace{-1.6ex}%
$ & constraint Jacobian before switching event \\ 
$\mathbf{J}_{+}$ & -$%
\hspace{-1.6ex}%
$ & constraint Jacobian after switching event \\ 
$\mathbf{F}$ & -$%
\hspace{-1.6ex}%
$ & orthogonal complement to $\mathbf{J}$ \\ 
$\mathbf{N}_{\mathbf{J},\mathbf{M}}$ & -$%
\hspace{-1.6ex}%
$ & projector to the null-space of $\mathbf{J}$ according to weight matrix $%
\mathbf{M}$ \\ 
$\mathbf{J}_{\mathbf{M}}^{+}$ & -$%
\hspace{-1.6ex}%
$ & weighted pseudoinverse of $\mathbf{J}$ with weight matrix $\mathbf{M}$
\\ 
$\mathbf{M}$ & -$%
\hspace{-1.6ex}%
$ & generalized mass matrix of the unconstrained system \\ 
$\mathbf{C}$ & -$%
\hspace{-1.6ex}%
$ & generalized Coriolis matrix of the unconstrained system \\ 
$\mathbf{P}$ & -$%
\hspace{-1.6ex}%
$ & vector of generalized potential forces of the unconstrained system \\ 
$\mathbf{Q}$ & -$%
\hspace{-1.6ex}%
$ & vector of remaining generalized forces of the unconstrained system \\ 
$\overline{\mathbf{M}}$ & -$%
\hspace{-1.6ex}%
$ & generalized mass matrix in terms of minimal coordinates \\ 
$\overline{\mathbf{C}}$ & -$%
\hspace{-1.6ex}%
$ & vector of generalized potential forces of the unconstrained system in
terms of minimal coordinates \\ 
$\overline{\mathbf{P}}$ & -$%
\hspace{-1.6ex}%
$ & vector of generalized potential forces of the unconstrained system in
terms of minimal coordinates \\ 
$\overline{\mathbf{Q}}$ & -$%
\hspace{-1.6ex}%
$ & vector of remaining generalized forces of the unconstrained system in
terms of minimal coordinates \\ 
$\mathbf{\lambda }$ & -$%
\hspace{-1.6ex}%
$ & vector of Lagrange multipliers \\ 
$\Lambda $ & -$%
\hspace{-1.6ex}%
$ & vector of impulsive forces \\ 
$\mathbf{u}$ & -$%
\hspace{-1.6ex}%
$ & generalized external (control) forces \\ 
$\mathbf{S}_{i}$ & -$%
\hspace{-1.6ex}%
$ & instantaneous screw coordinate vector of joint $i$ \\ 
$\mathbf{V}_{i}$ & -$%
\hspace{-1.6ex}%
$ & spatial twist vector of body $i$ \\ \hline
\end{tabular}

\end{document}